\documentclass[a4paper,fleqn]{cas-dc}
\usepackage[numbers]{natbib}
\usepackage{booktabs}
\usepackage{amsmath}
\usepackage{graphicx}
\usepackage[protrusion=true,expansion=false]{microtype}

\usepackage{xpatch}
\ExplSyntaxOn
\cs_set:Npn \__first_footerline:
  {
    \group_begin:
    \small \sffamily
    \ifnum\theblind>0\relax \else \__short_authors: :~ \fi
    { \rmfamily \itshape Preprint }
    \group_end:
  }
\ExplSyntaxOff

\begin{document}

\let\WriteBookmarks\relax
\def\floatpagepagefraction{1}
\def\textpagefraction{.001}
\renewcommand{\topfraction}{0.92}
\renewcommand{\bottomfraction}{0.75}
\renewcommand{\textfraction}{0.07}
\renewcommand{\floatpagefraction}{0.75}
\renewcommand{\dbltopfraction}{0.92}
\renewcommand{\dblfloatpagefraction}{0.72}
\setcounter{topnumber}{3}
\setcounter{bottomnumber}{2}
\setcounter{totalnumber}{4}
\setcounter{dbltopnumber}{3}

\shorttitle{What governs tooth segmentation on panoramic radiographs}
\shortauthors{M. Rehan et al.}

\title[mode=title]{Detection is solved, delineation is not: what governs tooth
segmentation on panoramic radiographs}

\author[1]{Muhammad Rehan}[orcid=0009-0008-9175-6714]
\cormark[1]
\ead{mrehan.msee21seecs@seecs.edu.pk}

\author[1]{Moaz Amjad}[orcid=0009-0007-3209-2922]
\ead{mamjad.msee21seecs@seecs.edu.pk}

\author[1]{Syed Danial Ahmed}[orcid=0009-0000-2849-6042]
\ead{syeddanialahmed@gmail.com}

\author[1]{Mariam Adnan}[orcid=0009-0001-0776-6792]
\ead{mariamadnan206@gmail.com}

\author[1]{Haider Ali}[orcid=0009-0004-2439-9447]
\ead{i171095@nu.edu.pk}

\affiliation[1]{organization={AI Dentify}, country={Pakistan}}
\cortext[1]{Corresponding author}

\begin{abstract}
Automatic tooth segmentation and FDI numbering on panoramic radiographs underpins
computer-assisted dental diagnosis, yet which factors govern performance remains unclear. We
assemble a corpus of 1{,}422 panoramic radiographs containing 42{,}142
expert-delineated tooth polygons across the 32-class FDI taxonomy, annotated by 30 dental
practitioners and independently reviewed by two others, and use it to isolate input
resolution, architecture and anatomical priors under a single evaluation protocol.

First, resolution dominates: across a controlled 640/1024/1280 ablation, mask mAP50-95 rises
$0.656 \rightarrow 0.710 \rightarrow 0.717$ while mAP50 stays flat at ${\sim}0.982$. Both
gains are significant under a paired bootstrap over images ($p < 0.001$, $p = 0.024$);
neither mAP50 change is distinguishable from zero. Added resolution buys boundary precision, not
detection. Second, architecture is nearly irrelevant in-domain: a query-based transformer with
$2.1\times$ the parameters is statistically equivalent to a one-stage detector
(95\% CI $[-0.0064, +0.0064]$), only marginally better under domain shift, $5.5\times$ slower on
CPU and not executable under standard ONNX runtimes. Third, three targeted interventions fail:
a LoRA-adapted self-supervised encoder underperforms, a promptable foundation segmenter
degrades masks by 39\%, and globally optimal anatomical label assignment yields $+0.0007$
despite correcting a constraint violated in 40\% of out-of-domain predictions.

Zero-shot transfer to an independent multi-centre cohort, verified overlap-free, costs
62\% of mask mAP50-95 but only 18\% of mAP50, reproducing the dissociation. Decomposing masks
along the tooth axis localises the residual error to the apical third. Boundary precision is
therefore the binding constraint, and effort is better directed at resolution and acquisition
diversity than at architectural novelty.
\end{abstract}

\begin{highlights}
\item Resolution, not architecture, governs accuracy: pixels buy boundaries, not detection
\item Error concentrates in the apical third: 0.78 IoU there against 0.90 at the crown
\item A 2.1x larger transformer ties a one-stage detector and will not run on ONNX CPU
\item SAM2 refinement degrades radiographic masks by 39 percent
\item Perceptual hashing fails as a contamination check on panoramic radiographs
\end{highlights}

\begin{keywords}
panoramic radiography \sep tooth segmentation \sep FDI numbering \sep
instance segmentation \sep domain shift \sep foundation models
\end{keywords}

\maketitle

\section{Introduction}
Panoramic radiographs (orthopantomograms, OPGs) are the most common extraoral dental
examination, and automatic delineation and numbering of individual teeth is a prerequisite for
downstream tasks such as charting, pathology localisation and periodontal measurement. Unlike
most medical segmentation problems, the task carries an unusual constraint: each structure must
be assigned an identity from a fixed anatomical taxonomy, the two-digit FDI notation, in which
position along the dental arch determines the label. Getting the mask right is not sufficient;
the model must also count correctly along an arch that may contain gaps.

Progress on this task is difficult to interpret. A recent systematic review and meta-analysis
of deep learning for tooth detection and segmentation on panoramic radiographs
\citep{dlreview} found that reported results typically come from a single train/validation
split of a single-centre corpus, and architectures are compared without holding resolution,
augmentation or post-processing fixed, so it is rarely clear whether a reported gain reflects
the model or the protocol surrounding it. Clinical evaluations of deployed systems report
markedly lower and more variable performance than development studies
\citep{aiaccuracy,aiplatform}, which is consistent with generalisation being the limiting
factor rather than in-domain accuracy. Two developments make this
more pressing. First, foundation models---self-supervised encoders and promptable
segmenters---are now routinely proposed for dental imaging, typically on the assumption that
large-scale pretraining transfers to radiography. Second, much of the field trains on
redistributed public corpora whose provenance and mutual overlap are undocumented
\citep{copycats}, so external validation may silently measure memorisation.

This paper asks a narrower question than ``which model is best'': \emph{what actually governs
performance on this task?} We hold the corpus, the evaluation code and the post-processing
constant, vary one factor at a time, and report every difference with a paired bootstrap
confidence interval so that effects can be judged rather than merely ranked---and so that a
claim of equivalence rests on a bounded interval rather than on a failure to reach
significance. Three of
the interventions we evaluate fail, and we regard these negative results as the paper's value
rather than its weakness: each rules out an approach a practitioner might otherwise reasonably
pursue, and each comes with an identified mechanism rather than a bare number.

Our contributions are:

\begin{enumerate}
  \item A demonstration that FDI numbering makes tooth instance segmentation equivalent to
        semantic segmentation, verified on all 1{,}422 images, which licenses encoder-only
        architectures and explains why class-agnostic promptable models cannot solve the task
        alone (Section~\ref{sec:uniqueness}).
  \item A controlled resolution ablation showing that resolution buys boundary precision and
        not detection, with paired bootstrap intervals establishing the mAP50-95 gain and the
        absence of any mAP50 gain (Section~\ref{sec:resolution}).
  \item An architecture comparison under identical metrics, including CPU latency and
        deployability, showing near-equivalence in-domain (Section~\ref{sec:architecture}).
  \item An anatomical decomposition of segmentation error showing it is not distributed over
        the tooth outline but concentrated in the apical third, where the root has no
        radiographic boundary (Section~\ref{sec:anatomy}).
  \item Three negative results, each with an identified mechanism
        (Section~\ref{sec:interventions}).
  \item A contamination-audit methodology for panoramic radiographs, and evidence that
        perceptual hashing is unreliable on this modality (Section~\ref{sec:audit}).
\end{enumerate}

\section{Related work}

\paragraph{Panoramic tooth segmentation corpora}
Several public corpora support tooth-level analysis of panoramic radiographs. The Tufts Dental
Database \citep{tufts} provides 1{,}000 radiographs with tooth masks, abnormality labels and
eye-tracking data. DENTEX \citep{dentex}, released for a MICCAI 2023 challenge, contributes
panoramic images from three institutions with quadrant and enumeration annotations, which
reconstruct FDI codes exactly. AKUDENTAL \citep{akudental} provides 333 images with 9{,}956
annotated structures across 32 FDI classes plus three restorative categories, and is the
closest comparator to the present work in taxonomy and intent. Larger corpora exist for
adjacent tasks: a dual-labelled set of 5{,}000 radiographs (2{,}066 public) adds tooth state
categories \citep{duallabelled}, and OralXrays-9 \citep{oralxrays} provides 12{,}688 images
annotated for nine anomaly categories rather than tooth identity. Systematic review of
oral-maxillofacial datasets \citep{npjreview} found that 83.8\% do not disclose whether ethical
approval was obtained and only 25.7\% state annotator qualifications, which motivates the
protocol reporting in Section~\ref{sec:annotation}.

\paragraph{Tooth segmentation and numbering}
Prior work on panoramic tooth analysis spans supervised instance segmentation, self-supervised
pretraining for tooth numbering and restoration detection \citep{mim_teeth}, and
semi-supervised approaches evaluated through the STS challenge \citep{sts}. Several studies
combine multiple public corpora to increase scale \citep{combinepublic}, and multinational
evaluation across three regions reports substantial variation in performance by site
\citep{multinational}. Direct adaptation of promptable foundation models to this task has also
been attempted \citep{sam2tooth}; we revisit that approach in
Section~\ref{sec:interventions} and reach a different conclusion.

\paragraph{Foundation models in medical segmentation}
The Segment Anything Model \citep{sam} and its successor SAM2 \citep{sam2} established
promptable, class-agnostic segmentation trained at web scale, and medical adaptations followed
rapidly \citep{medsam}. Self-supervised encoders have developed in parallel \citep{dinov2}: DINOv3
\citep{dinov3} is pretrained on 1.7B images and produces dense features that transfer without
domain-specific pretraining. DinoDental \citep{dinodental} benchmarks DINOv3 across dental
classification, detection and segmentation tasks and reports competitive results against
supervised backbones, comparing frozen features, full fine-tuning and LoRA adaptation. Both
families have been proposed for dental imaging; we evaluate both here and find neither improves
on a conventional detector (Sections~\ref{sec:architecture} and~\ref{sec:interventions}).

\paragraph{Instance segmentation architectures}
Encoder--decoder convolutional networks remain the default in medical segmentation
\citep{unet,deeplabv3p}, while two families dominate instance segmentation: region-based
detectors \citep{maskrcnn} and, more recently, query-based transformers. Mask2Former
\citep{mask2former} predicts masks directly through masked attention over learned queries,
built on a hierarchical vision backbone \citep{swin}, and is the stronger family on general
benchmarks; the YOLO series \citep{ultralytics} remains preferred where inference cost
matters. We compare both under an identical
protocol and additionally report deployability, which is rarely stated but frequently decisive.

\paragraph{Configuration versus architecture}
Our central finding has a precedent outside dentistry. nnU-Net \citep{nnunet} showed that a
systematically configured but architecturally plain network matches or beats bespoke designs
across a wide range of biomedical segmentation tasks, arguing that preprocessing, resolution
and training schedule dominate architectural choice. Related work on distribution shift shows
that robustness under acquisition change is likewise governed less by architecture than by the
data-generating process \citep{distshift,modelsoup}. We reach a similar conclusion for
panoramic tooth segmentation, and additionally quantify the effect against cross-validated
variance.

\paragraph{Dataset redistribution and contamination}
Public medical imaging corpora proliferate uncontrolled across community platforms.
\citet{copycats} document 640 ISIC-derived datasets on one platform against a 38\,GB official
release, 24 separate uploads of INBreast, and widespread loss of licence and provenance
metadata, together with patient-level duplication that leaks between training and test
partitions. The consequences are not hypothetical: a redistributed medical dataset was withdrawn
for copyright infringement in 2026, with associated retractions. This motivates the
contamination audit in Section~\ref{sec:audit} and our finding that the usual perceptual-hash
check is unreliable on panoramic radiographs.

\section{Materials and methods}

\subsection{Corpus}
\label{sec:corpus}
\begin{figure*}[pos=t]\centering
\includegraphics[width=\linewidth]{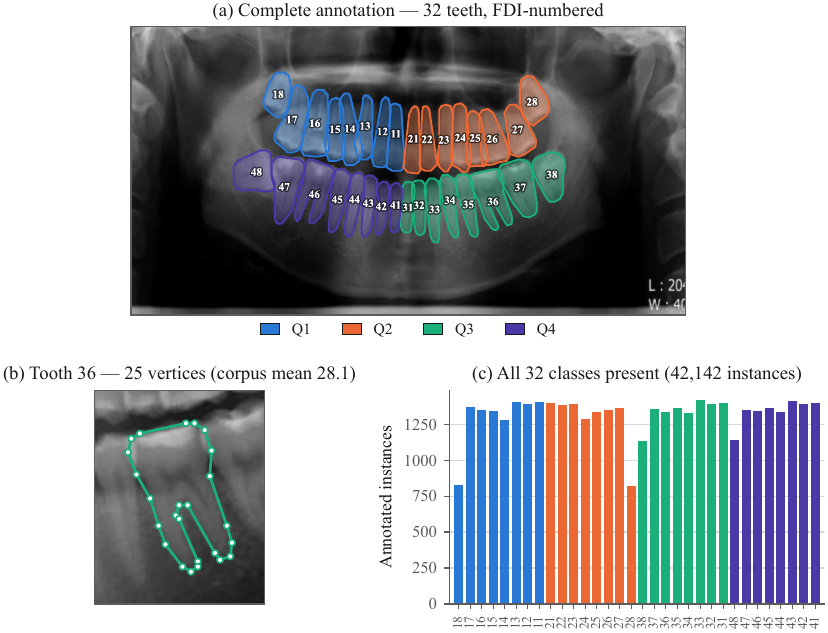}
\caption{The annotation layer. (a) A complete dentition, coloured by FDI quadrant and
numbered; fill colour encodes quadrant, a property of the taxonomy rather than arbitrary
identity. (b) Polygon granularity for a tooth at the corpus mean vertex count; the bifurcated
roots of the first molar are traced individually. (c) Per-class instance counts: all 32 classes
are represented, with third molars the only markedly less frequent group.}
\label{fig:dataset}
\end{figure*}

The corpus comprises 1{,}422 panoramic radiographs with 42{,}142 tooth polygons, a mean of
29.6 annotated teeth per image, with all 32 FDI classes present in both partitions. Annotations
were produced by 30 dental practitioners at 28.1 vertices per tooth and independently reviewed
by two reviewing dentists (Section~\ref{sec:annotation}).
Figure~\ref{fig:dataset} shows a representative annotation, the polygon granularity, and the
per-class instance distribution. Polygons contain 28.1 vertices per tooth on average (median
25), against 15.9 points per instance reported for AKUDENTAL \citep{akudental}; multi-rooted
teeth are traced root by root. Class frequencies are near-uniform apart from the third molars
(18, 28, 38, 48), which occur in roughly 820--1{,}140 images against 1{,}250--1{,}420 for other
classes. That imbalance reflects extraction and agenesis rather than annotation bias, and is
therefore the true prior; no resampling was applied.

Radiographs were obtained from a publicly available redistribution; the upstream clinical
origin is not documented by the redistributor and we therefore make no claim to the imagery.
Our contribution is the annotation layer. Section~\ref{sec:audit} establishes that the corpus
does not overlap the datasets used for external validation.

The redistributed files were not de-identified: filenames encoded patient names and acquisition
timestamps. We stripped them and reassigned sequential identifiers before annotation began, so
the corpus described here contains no identifier and no annotator ever saw one; the mapping is
held offline and not released. This is a concrete instance of the provenance failure documented
for redistributed medical corpora \citep{copycats}, and it is a further reason the imagery
cannot be redistributed by us.

Duplicate images were removed in three passes: exact MD5, perceptual hash, and exhaustive
correlation of brightness- and contrast-normalised thumbnails. The third pass identified a
pixel-identical pair that the first two missed, and a further pair
($r{=}0.962$, 7-pixel offset, no monotone intensity relation) consistent with one patient
imaged twice; the latter was constrained to a single cross-validation fold.

\subsection{Annotation protocol}
\label{sec:annotation}
Annotations were produced by 30 dental practitioners over 218 days and independently reviewed
by two reviewing dentists. All annotators hold a Bachelor of Dental Surgery (BDS) with varying
clinical experience; the two reviewers hold a BDS with four and three years of clinical
experience respectively. Every included image passed review: of 1{,}491 approved annotations,
none lacks a recorded reviewer, and no reviewer ever approved an image they had annotated
themselves. Median review turnaround was 32.5\,h. Polygons contain 28.1 vertices per tooth on
average (median 25).

No automatic pre-annotation was used: every polygon was drawn manually from scratch, so the
labels are independent of any segmentation model and carry no risk of circularity when used to
evaluate one. Annotation guidance was fixed for the full 218-day period; no rules were revised
mid-project, so no subset of the corpus was annotated under different conventions.

\paragraph{Conventions applied, and their consistency}
No written boundary protocol was issued: annotators worked to conventions established during
training. Rather than reconstruct those conventions from recollection seven months later, we
characterise them from the polygons. Masks include the root as well as the crown (median
elongation, axial over transverse extent, $3.30$; a crown-only outline would be $1.0$--$1.4$).
Teeth were outlined independently rather than as a partition of the image, so adjacent teeth may
share pixels where they overlap in projection: $4.87\%$ of same-image class pairs overlap, by a
median $2.1\%$ of the smaller mask. Teeth were annotated within the frame, with $8$ of $11{,}974$
sampled polygons touching an image border.

The apical boundary is the least determinate of these, and the one our results turn on
(Section~\ref{sec:anatomy}), so we tested whether it was applied consistently. Elongation is a
ratio, so it is invariant to patient size and image scale, and an annotator who traces roots
further shows systematically higher values. Centring within (class, track) cells to remove tooth
shape and case mix, and comparing the 22 annotators who contributed at least 200 teeth
($41{,}494$ teeth in total), one-way random-effects ANOVA gives an intraclass correlation of
$0.030$: $3\%$ of residual variation is attributable to annotator identity and $97\%$ to
anatomy within annotators. Most annotators sit within $\pm2.5\%$ of the consensus root extent.
One is a clear outlier at $+10.3\%$ (95\% CI $[+9.4, +11.8]$), and $2.5\%$ of the corpus comes
from annotators deviating by $5\%$ or more.

This is not inter-annotator agreement, and we do not present it as a substitute: no two
annotators traced the same tooth, so per-image disagreement remains unmeasured. What it bounds is
the systematic component---whether some practitioners drew consistently longer roots than
others---which is the failure mode a 30-annotator corpus most invites, and it is small.

Primary and supernumerary teeth were excluded by protocol. This exclusion is load-bearing for
Section~\ref{sec:uniqueness}.

\subsection{FDI uniqueness reduces instance to semantic segmentation}
\label{sec:uniqueness}
FDI notation assigns each permanent tooth a unique identifier within a mouth, so a given class
can appear at most once per image. We verified this on all 1{,}422 annotated images: no image
contains a repeated class.

Consequently 32-class instance segmentation is equivalent to 33-class semantic segmentation
(32 teeth plus background), and each predicted class map \emph{is} its own instance. No
detection stage, mask proposals or non-maximum suppression are required. This is what makes
encoder-only architectures directly applicable, and it identifies why promptable
class-agnostic models such as SAM cannot solve the task alone: they produce masks but not
identities, and identity is the difficult part. The property fails for supernumerary teeth
(approximately 1--3\% of patients), which this protocol excludes.

\subsection{Contamination audit}
\label{sec:audit}
Because our imagery derives from a public redistribution, no external corpus can be assumed
free of overlap; evaluating on a contaminated set would measure memorisation rather than
generalisation. Filenames are uninformative---redistributions routinely renumber---so comparison
must be on pixel content.

All training images were compared against the Tufts Dental Database ($n{=}1{,}000$) and DENTEX
($n{=}3{,}904$) by MD5, perceptual hash, and exhaustive all-pairs correlation of
brightness/contrast-normalised embeddings including mirrored variants (1.42M comparisons).
No overlap was found: maximum correlation to any Tufts image was $0.857$, against $1.000$ for
verified duplicates, and zero MD5 matches to either dataset.

Perceptual hashing proved unreliable on this modality. The median nearest-neighbour perceptual
hash distance from our corpus to Tufts was 12 bits, identical to the median distance
\emph{within} our own set of distinct radiographs, because all panoramic images share the same
global arch structure. Thresholds calibrated on natural images therefore produce false
positives here, and exact hashing plus pixel correlation is required.

\subsection{Models and training}
\label{sec:models}
We evaluate three architectures spanning the dominant families. \textbf{YOLO11m-seg} \citep{ultralytics} (22.4M
parameters) is a one-stage detector with a prototype-based mask head. \textbf{Mask2Former}
\citep{mask2former} with a Swin-tiny backbone \citep{swin} (47.4M) represents query-based transformers; note
that it carries $2.1\times$ the parameters of the detector, so the comparison favours it on
capacity. \textbf{DINOv3-B} \citep{dinov3} (93.0M) is used as a frozen self-supervised encoder
adapted with LoRA \citep{lora} on the attention projections (7.8\% of parameters trainable),
with a DPT-style decoder \citep{dpt} fusing four encoder depths; by Section~\ref{sec:uniqueness} a dense 33-class
classifier is sufficient, so no detection head is required.

All models were trained at $1280 \times 672$, preserving the $1.92{:}1$ aspect ratio of the
source radiographs. This matters for efficiency as well as accuracy: a rectangular canvas of
$1280 \times 672$ contains 860k pixels against 1{,}049k for a $1024^2$ square, so it provides
greater resolution along the dental arch while performing less computation, because square
letterboxing spends roughly a third of the frame on padding.

Augmentation is deliberately photometric. Brightness, contrast, gamma and additive noise are
applied, since acquisition variation across scanners is the dominant source of domain shift.
Geometric augmentation is minimal because panoramic acquisition geometry is standardised, and
\textbf{horizontal flipping is disabled entirely}: mirroring the image exchanges the left and
right quadrants, converting tooth 18 into tooth 28 and inverting every FDI label. Standard
detection pipelines enable horizontal flip by default, which would silently corrupt a large
fraction of labels on this task.

Optimisation used AdamW \citep{adamw} with a cosine schedule \citep{sgdr}. Class imbalance is mild (third molars appear
approximately $830$ times against $1{,}400$ for central incisors, a $1.7{:}1$ ratio) and
reflects genuine extraction and agenesis rather than annotation bias, so no resampling was
applied.

\subsection{Evaluation}
All models are scored with identical code on identical mask representations, at original image
resolution. Mask mAP follows the COCO protocol \citep{cocometric} and is computed with
\texttt{pycocotools}; mIoU is accumulated dataset-wide
per class. One instance per FDI class is retained for every model, so comparisons isolate the
architecture rather than the post-processing.

\subsection{Statistical analysis}
\label{sec:stats}
Differences between models are assessed by paired bootstrap over images. Both models in a
comparison are scored on the same resampled image set, which preserves the correlation induced
by shared images and shared difficult cases and gives a tighter interval than an unpaired
resample. We draw $2{,}000$ resamples and report the observed difference, a 95\% percentile
confidence interval, and a two-sided bootstrap $p$-value. Within each metric, $p$-values are
Holm-corrected across the family of comparisons reported here.

Recomputing mAP on a resample is exact rather than approximate. FDI uniqueness (Section
\ref{sec:uniqueness}) guarantees at most one ground-truth instance per class per image, and the
post-processing enforces at most one detection, so every (image, class) cell holds at most one
of each and COCO's greedy matching has nothing to disambiguate. Caching per-cell detection
scores and IoUs therefore permits exact reconstruction of the COCO 101-point interpolation for
any subset of images. We verified that the reconstruction reproduces the evaluation harness on
the full set to within $5\times10^{-5}$ for every model and metric.

Two questions are kept separate throughout, because a large $p$-value is not evidence of
equivalence: whether a difference exists is answered by the interval excluding zero, and whether
two models are interchangeable is answered by the interval lying wholly inside a margin. For the
latter we report the equivalence margin---the smallest symmetric bound containing the whole
confidence interval---so the reader can judge it against a threshold they consider clinically
negligible rather than against an arbitrary $\alpha$.

\section{Results}

\subsection{Resolution}
\label{sec:resolution}
\begin{figure}[pos=t]\centering
\includegraphics[width=\linewidth]{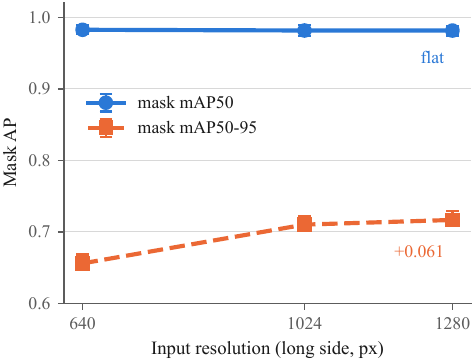}
\caption{Input resolution acts on boundary precision, not detection. mAP50 is flat to three
decimal places across the range while mAP50-95 rises by $0.061$. Error bars are 95\% bootstrap
intervals over images; the mAP50 intervals overlap almost entirely.}
\label{fig:resolution}
\end{figure}

Table~\ref{tab:resolution} and Fig.~\ref{fig:resolution} report the ablation. Mask mAP50-95
rises monotonically while mAP50 is constant to three decimal places.

\begin{table}[pos=t]\centering
\caption{Input resolution ablation, identical protocol. $\Delta$ is the change in mask mAP50-95
from the preceding row, with a 95\% paired bootstrap interval over images. Best epoch rises
$29 \rightarrow 40 \rightarrow 81$ across the three arms.}
\label{tab:resolution}
\begin{tabular}{lccc}
\toprule
Input & mAP50 & mAP50-95 & $\Delta$ mAP50-95 \\
\midrule
640  & 0.983 & 0.656 & --- \\
1024 & 0.982 & 0.710 & $+0.054$ \\
     &       &       & \scriptsize$[+0.048, +0.060]$ \\
1280 & 0.982 & \textbf{0.717} & $+0.007$ \\
     &       &       & \scriptsize$[+0.002, +0.011]$ \\
\bottomrule
\end{tabular}
\end{table}

Both resolution increases improve mAP50-95 significantly ($p < 0.001$ and $p = 0.024$
respectively, Holm-corrected), and neither improves mAP50: the $640 \rightarrow 1024$ change in
mAP50 is $-0.001$ (95\% CI $[-0.005, +0.003]$) and the $1024 \rightarrow 1280$ change is
$-0.000$ (95\% CI $[-0.004, +0.004]$). The dissociation is therefore not an artefact of reading
two curves by eye: the same intervention that reliably moves boundary quality is
statistically indistinguishable from no change at the detection threshold.

Five-fold grouped, track-stratified cross-validation gives mask mAP50-95
$0.7216 \pm 0.0035$ (range $0.7176$--$0.7254$) under the training framework's own validator.
Notably, mAP50 varies less across folds ($\pm 0.0021$) than mAP50-95 ($\pm 0.0035$): the
variance structure reproduces the same dissociation.

Best epoch increases with resolution ($29 \rightarrow 40 \rightarrow 81$), so a fixed
early-stopping patience systematically understates higher-resolution arms.

\subsection{Architecture}
\label{sec:architecture}
\begin{table*}[pos=t]\centering
\caption{Architectures under an identical evaluation protocol. The in-domain difference between
the top two is $0.0003$ (95\% CI $[-0.0064, +0.0064]$): not merely non-significant, but bounded
inside $\pm0.007$ mAP.}
\label{tab:arch}
\begin{tabular}{lrrrrl}
\toprule
Model & Params & In-domain mAP50-95 & Zero-shot mAP50-95 & CPU (ms) & ONNX \\
\midrule
YOLO11m-seg          & 22.4M & \textbf{0.7166} & 0.2740 & \textbf{749}  & yes \\
Mask2Former swin-T   & 47.4M & 0.7163 & \textbf{0.2823} & 4105 & \textbf{no} \\
DINOv3-B + LoRA      & 93.0M & 0.6470 & 0.2309 & --- & --- \\
\bottomrule
\end{tabular}
\end{table*}

Table~\ref{tab:arch} reports the three architectures under the common protocol, in-domain and
zero-shot, alongside CPU latency and deployability.

The in-domain comparison supports a stronger statement than ``no significant difference''. The
paired bootstrap places the YOLO11m--Mask2Former gap at $+0.0003$ mask mAP50-95 with a 95\%
interval of $[-0.0064, +0.0064]$, so the two are equivalent within $\pm0.007$ mAP---roughly a
tenth of what raising input resolution from 640 to 1024 delivers on the same data. Neither of
the other two metrics separates them either: mAP50 differs by $+0.0024$
(CI $[-0.0016, +0.0068]$) and mIoU by $-0.0018$ (CI $[-0.0069, +0.0037]$). Because the interval
is narrow rather than merely straddling zero, this is a claim about equivalence and not an
artefact of an underpowered test.

Out of domain the picture changes, and the change is worth stating precisely rather than
folding into the same conclusion. On DENTEX, Mask2Former is better on every metric and every
difference is significant after correction: mask mAP50-95 $+0.0083$ (95\% CI
$[+0.0039, +0.0122]$), mAP50 $+0.0323$ ($[+0.0234, +0.0401]$), mIoU $+0.0164$
($[+0.0108, +0.0221]$). Architecture is therefore irrelevant in-domain and modestly but reliably
relevant under domain shift---a distinction that a single in-domain comparison would have
hidden entirely, and one that argues the benefit of a stronger architecture is a generalisation
benefit rather than a fitting benefit.

Mask2Former could not be served: its deformable attention lowers to \texttt{GridSample}, which
the ONNX Runtime CPU execution provider does not implement at opset 17 or 20. The graph exports
without error and fails at session creation.

\subsection{Cross-dataset transfer}
\begin{figure}[pos=t]\centering
\includegraphics[width=\linewidth]{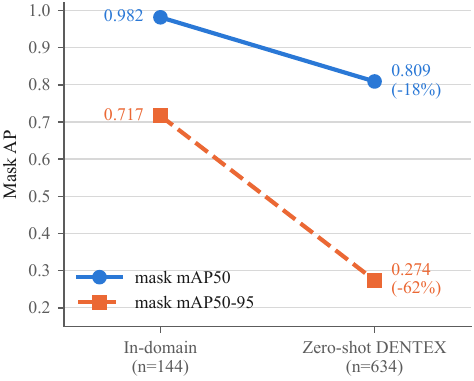}
\caption{Domain shift reproduces the dissociation from a second direction. Zero-shot transfer
to DENTEX costs 62\% of mask mAP50-95 but only 18\% of mAP50.}
\label{fig:transfer}
\end{figure}

Zero-shot evaluation on 634 DENTEX images (FDI reconstructed as
$\text{quadrant} \times 10 + \text{position}$, an exact match to our taxonomy) costs 62\% of
mask mAP50-95 ($0.7166 \rightarrow 0.2740$) but only 18\% of mAP50
($0.9815 \rightarrow 0.8089$). The model continues to find and correctly number roughly 81\% of
teeth on unseen equipment; what degrades is where it places the boundary
(Fig.~\ref{fig:transfer}).

\begin{figure}[pos=t]\centering
\includegraphics[width=\linewidth]{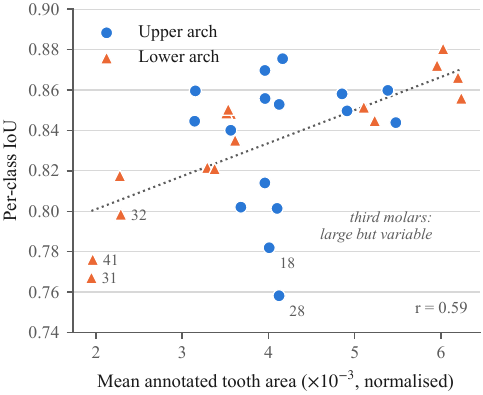}
\caption{Per-class IoU against mean annotated tooth area ($r{=}0.59$). The smallest teeth
(lower incisors 41, 31, 32) are weakest. Third molars (18, 28) are a documented exception:
large, but frequently impacted or partially erupted.}
\label{fig:size}
\end{figure}

\subsection{Where on the tooth the error lives}
\label{sec:anatomy}
The preceding sections establish that boundary precision is the binding constraint but not
which part of the boundary is responsible. We localise it anatomically. For every correctly
detected tooth, principal component analysis of the ground-truth mask gives the long axis;
pixels are projected onto it and divided into three equal bands from crown to apex, with
crown/apex orientation resolved per quadrant because the upper and lower arches point in
opposite directions in a panoramic projection. IoU is then computed within each band.

\begin{table}[pos=t]\centering
\caption{Segmentation accuracy by position along the tooth axis, over 4{,}207 detected teeth.
Error is concentrated in the apical third.}
\label{tab:anatomy}
\begin{tabular}{lcc}
\toprule
Band & IoU & Share of union area \\
\midrule
Crown third  & 0.905 & 39.5\% \\
Middle third & 0.906 & 35.2\% \\
Apical third & \textbf{0.776} & 25.4\% \\
\bottomrule
\end{tabular}
\end{table}

\begin{figure*}[pos=t]\centering
\includegraphics[width=\textwidth]{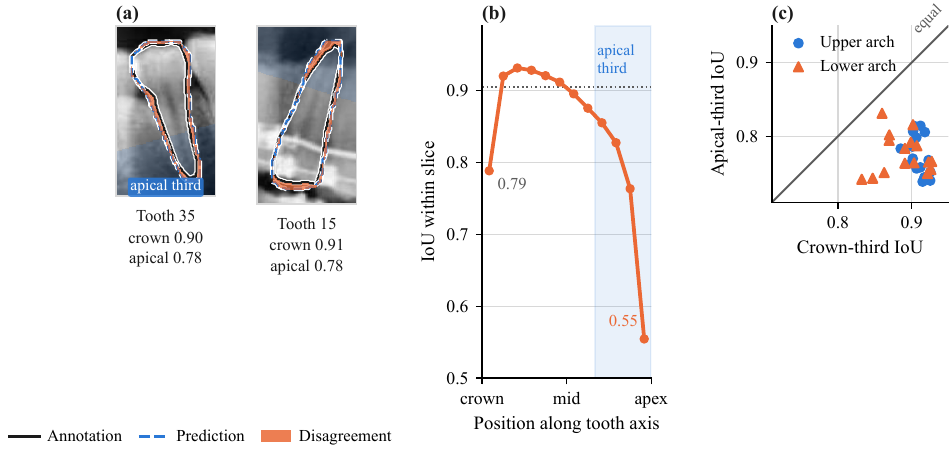}
\caption{Segmentation error is concentrated at the root apex. (a) Two teeth with the annotation,
the prediction, and their disagreement; the apical third is tinted. Panels are selected
automatically as the instances whose crown and apical IoU lie closest to the corpus means, so
they are typical rather than extreme. (b) IoU against position along the tooth axis, pooled over
4{,}208 teeth. Both tips fall away---thin end slices are unforgiving---but only the apical side
keeps falling, reaching $0.55$ in the final slice against $0.79$ at the occlusal tip. (c) Every
FDI class lies below the diagonal, so the effect is not driven by a subset of classes.}
\label{fig:anatomy}
\end{figure*}

The result is unambiguous (Table~\ref{tab:anatomy}, Fig.~\ref{fig:anatomy}). The crown and middle
thirds are delineated almost identically well, and accuracy collapses by roughly 13 points in the
apical third. The displacement of the predicted mask centroid decomposes the same way: $3.79$ px
along the tooth axis against $1.01$ px across it, a ratio of $3.8$. The model recovers the width
of a tooth accurately and its length poorly.

Resolving the axis into finer slices (Fig.~\ref{fig:anatomy}b) shows the decline is not a
symmetric endpoint artefact, which is the obvious alternative explanation: a thin slice at either
tip contains few pixels, so a fixed boundary offset costs proportionally more IoU there. Both
ends do dip. But the occlusal tip dips once, to $0.79$, and recovers to $0.93$ in the next slice,
whereas the apical side declines monotonically across the final four slices to $0.55$. Comparing
slices equidistant from each tip makes the asymmetry explicit ($0.79$ against $0.55$ at the
extreme, $0.92$ against $0.76$ one slice in), so geometry alone does not account for it.

This localises the residual error to exactly the structure that has no radiographic edge. Enamel
and dentine present genuine contrast; the root apex fades into trabecular bone, and where the
apex ends is a judgement rather than an observation. Three findings converge on that same
region: the base model loses most of its accuracy there, SAM2 fails there catastrophically
rather than uniformly (Section~\ref{sec:interventions}), and multi-rooted molars (16, 26, 46;
apical IoU $0.740$--$0.749$) and thin-rooted lower incisors (31, 41; $0.742$, $0.743$) are worst
of all, while single-rooted teeth with well-defined apices (38, 35, 23, 13) are best at
$0.814$--$0.831$.

The practical reading is that the remaining headroom on this task is not distributed over the
tooth outline but concentrated in one anatomically identifiable region, and that a share of it
may not be model error at all: the apical third is also where an annotation protocol is least
determinate, so part of this gap plausibly reflects the ceiling imposed by label ambiguity
rather than a deficiency the model could remove. Separating those two would require repeat
annotation at the apex, which we did not perform.

\subsection{Interventions that fail}
\label{sec:interventions}
\begin{table}[pos=t]\centering
\caption{Targeted interventions, change in mask mAP50-95 with 95\% paired bootstrap intervals.
Hungarian assignment is statistically detectable out of domain and practically negligible.}
\label{tab:interventions}
\setlength{\tabcolsep}{4pt}
\begin{tabular}{lcc}
\toprule
Intervention & In-domain & Zero-shot \\
\midrule
SAM2 refinement & $-0.278$ & $-0.054$ \\
  & \scriptsize$[-0.289, -0.262]$ & \scriptsize$[-0.057, -0.050]$ \\
\addlinespace
Hungarian assignment & $+0.0001$ & $+0.0007$ \\
  & \scriptsize$[-0.0000, +0.0003]$ & \scriptsize$[+0.0001, +0.0011]$ \\
\bottomrule
\end{tabular}
\end{table}

Table~\ref{tab:interventions} reports the two post-hoc interventions, both applied to the same
detector output. The third failed intervention, the LoRA-adapted self-supervised encoder, is a
trained architecture rather than a post-processing step and so appears in
Table~\ref{tab:arch}.

\paragraph{Promptable boundary refinement}
Since the preceding results identify boundary precision as the binding constraint, a promptable
segmenter with strong edge priors is the natural remedy, and adaptations of SAM2 to tooth
segmentation have been proposed \citep{sam2tooth}. We prompted SAM2 \citep{sam2} with each
detection's bounding box and selected, among the returned candidates, the mask with highest
overlap against the detector's original mask---the detector determines \emph{which} object,
SAM2 only redraws its outline. A refined mask was accepted only if it agreed with the original
above $\mathrm{IoU}=0.5$, making the procedure conservative by construction.

\begin{figure*}[pos=t]\centering
\includegraphics[width=\linewidth]{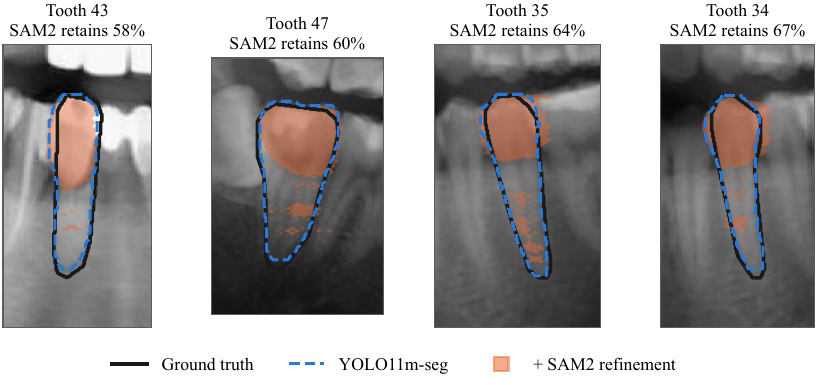}
\caption{Why promptable refinement fails on radiographs. Panels are selected automatically as
the teeth with the largest area reduction under refinement, one per class. SAM2's region
(orange) covers the crown and stops at the cervical margin, while ground truth (black) and the
detector's mask (blue, dashed) continue along the root. Retained area is 58--67\% of the
detector's mask.}
\label{fig:sam2}
\end{figure*}

The result is strongly negative. SAM2 replaced 98.1\% of masks and degraded mask mAP50-95 from
$0.7166$ to $0.4391$, a 39\% relative reduction. The signature identifies the mechanism: mAP50
fell only $0.026$ while mAP50-95 fell $0.278$, so the refined masks still cover the correct
tooth but place its edge incorrectly. Refined masks were consistently tighter than the annotation, and
Fig.~\ref{fig:sam2} shows the loss is concentrated in the root rather than spread evenly
around the outline. SAM2's prior is learned from natural images, where objects have genuine
intensity boundaries. A tooth in a panoramic radiograph does not: the cervical margin fades into
gingiva, the root into trabecular bone, and adjacent teeth overlap in projection. Faced with
that ambiguity SAM2 snaps to the high-contrast enamel, which lies inside the annotated outline.

Figure~\ref{fig:sam2} shows the mechanism directly. Across the most severely affected teeth,
SAM2 retains only 58--67\% of the mask area, and the pattern is identical in each: the refined
region covers the crown and terminates at the cervical margin, below which it produces only
scattered fragments. The failure is therefore not uniform shrinkage but a loss of the root
entirely. Enamel presents genuine radiographic contrast; the root fades into trabecular bone
with no edge for a natural-image prior to attach to.

This also exposes a limitation of agreement-based acceptance gates. Requiring overlap with the
original mask verifies that the refiner found the same \emph{object}, which 98\% of masks did,
but is blind to a uniform directional bias in where the edge is placed. Raising the threshold
does not help, because the error is systematic rather than sporadic.

\paragraph{Anatomical label priors}
FDI uniqueness (Section~\ref{sec:uniqueness}) is violated by ground truth in $0.00\%$ of the
1{,}422 annotated images, but by predictions in $9.7\%$ of in-domain and $39.9\%$ of zero-shot
images---a provable error in a large fraction of out-of-domain output, and therefore apparently
substantial headroom. We formulated assignment as a linear assignment problem over predicted
regions and the 32 FDI slots, solved optimally by the Hungarian algorithm \citep{hungarian}, with per-class
position priors learned from the training split. Those priors are strong: tooth centroids
occupy highly repeatable image coordinates ($x$ standard deviation $0.013$--$0.042$), and only
10 of 120 class pairs overlap at one standard deviation, so position alone nearly identifies a
tooth. Passing all 4{,}240 ground-truth instances through the solver returned them unchanged,
confirming it cannot damage correct predictions.

It gains $+0.0007$ mask mAP50-95 zero-shot (95\% CI $[+0.0001, +0.0011]$). The interval
excludes zero, so the gain is real rather than noise; it is also two orders of magnitude smaller
than the resolution effect, and it comes at a small but equally significant cost in mIoU
($-0.0037$, $[-0.0048, -0.0025]$). Statistical detectability and practical relevance part company
here, which is precisely why we report intervals rather than significance alone. Two factors explain this. Greedy
selection by confidence already resolves duplicates correctly in most cases, because the
confidence gap between a duplicated pair is wide (median $0.225$). More fundamentally, the
residual error was never in label space: zero-shot mAP50 remains $0.809$ while mAP50-95 falls
to $0.274$, so the model identifies teeth correctly and draws them loosely. An intervention
targeting identity cannot recover error located in geometry.

We also evaluated whether the whole-arch left/right flip check used during annotation review
transfers to model output. It does not fire once in 778 predicted images: annotators
occasionally invert an arch, models never do. We report this because it illustrates that
constraints useful for auditing human annotation do not automatically transfer to auditing
model predictions.

\section{Discussion}

\paragraph{One constraint, five independent lines of evidence}
Our results converge on a single mechanism. Raising input resolution moves mask mAP50-95 by
$0.061$ while leaving mAP50 flat at ${\sim}0.982$, a change indistinguishable from zero at
every step. Transferring to an unseen multi-centre cohort
destroys 62\% of mAP50-95 but only 18\% of mAP50. Per-class accuracy correlates with tooth
size ($r{=}0.59$, Fig.~\ref{fig:size}), the smallest teeth being weakest under every
architecture and resolution tested. An intervention aimed squarely at label space buys a
gain two orders of magnitude smaller than the resolution effect, despite correcting a constraint
violated in 40\% of out-of-domain predictions. And decomposing the mask along the tooth axis
locates the residual error in the apical third (Section~\ref{sec:anatomy}), where the structure
has no radiographic edge---the same region where promptable refinement fails outright.

The size relationship is real but partial, and the exception is informative. The third molars
(18, 28) are among the largest annotated structures yet rank with the lower incisors among the
weakest classes. Unlike the incisors, whose difficulty is one of scale and crowding, third
molars are frequently impacted, rotated or only partially erupted, so their boundary is
ambiguous for a different reason. Both failure modes are boundary-placement problems; neither
is a failure to detect or identify the tooth.

The fourth of these is the most informative, because it is a failure. The first three are
mutually consistent observations, and consistent observations can share a confound. A failed
intervention instead \emph{excludes} an alternative account: had a meaningful share of the error
been misassignment of identity, optimal assignment would have recovered it. It did not.
Detection and identification are effectively solved on this task at any resolution we tested;
what remains difficult, and what degrades under domain shift, is placing the boundary.

\paragraph{Architecture matters less than expected}
A query-based transformer with $2.1\times$ the parameters is statistically equivalent to a
one-stage detector in-domain ($0.7163$ vs $0.7166$; 95\% CI $[-0.0064, +0.0064]$) and is better
by $0.0083$ under domain shift (95\% CI $[+0.0039, +0.0122]$). That difference is real but
small---an eighth of the $0.061$ gained by moving from 640 to 1280 pixels. For
practitioners the implication is direct: on this task, effort spent on input resolution and
acquisition diversity is better rewarded than effort spent selecting an architecture. This
echoes the conclusion of nnU-Net \citep{nnunet} in a different domain; we quantify it here
against paired bootstrap intervals, and bound the architecture effect rather than merely failing
to detect it.

By contrast, both foundation-model approaches underperformed. A LoRA-adapted self-supervised
encoder trailed on every metric, and its deficit was roughly twice as large on mAP50-95 as on
mIoU, consistent with a decoder that bilinearly upsamples a coarse patch grid and therefore
produces systematically softer boundaries. A promptable segmenter actively degraded results.
Our zero-shot result is also consistent with the site-to-site variation reported in
multinational evaluation of dental AI \citep{multinational}. We do not read these findings as
evidence that foundation models are unsuitable for medical imaging in general, but they are a concrete counterexample to the assumption that web-scale pretraining
transfers usefully to radiographic boundary delineation, where the visual statistics that make
such priors work are largely absent.

\paragraph{Deployability can outrank accuracy}
Mask2Former cannot be served through ONNX Runtime's CPU execution provider: its deformable
attention lowers to \texttt{GridSample}, unimplemented at opset 17 and 20, so the graph exports
without error and fails at session creation. It is also $5.5\times$ slower on CPU
($4105$\,ms vs $749$\,ms). For a system performing CPU inference, that combination outweighs a
$0.008$ mAP advantage. Architecture comparisons in the clinical literature seldom report
whether the recommended model can actually be deployed on the target runtime; we suggest this
belongs alongside accuracy whenever a paper makes a deployment claim.

\paragraph{Contamination auditing needs modality-specific methods}
Perceptual hashing is the standard tool for detecting duplicated images between corpora, and it
fails here. The median nearest-neighbour perceptual-hash distance from our corpus to Tufts
equals the median distance \emph{within} our own set of distinct radiographs, because every
panoramic image shares the same global arch structure. A threshold calibrated on natural images
will therefore either miss true duplicates or flag unrelated patients. Given how much dental
work now trains on redistributed corpora \citep{copycats}, we recommend exact hashing combined
with exhaustive normalised pixel correlation, and reporting the measured separation rather than
a bare threshold.

\section{Limitations}
Several limitations bound these conclusions.

The imagery derives from a public redistribution whose upstream clinical origin is not
documented, so patient demographics, acquisition equipment and case mix are unknown, and the
corpus should be treated as effectively single-source. This is also why the paper claims only
the annotation layer.

Each image was annotated by one practitioner, so inter-annotator agreement could not be
computed and we cannot establish the human ceiling for this task. This is a material gap: it
prevents us from determining whether the residual error on lower anterior teeth reflects model
limitation or genuine clinical ambiguity. We report the two-stage independent review protocol in
its place, together with a direct measurement of between-annotator consistency
(Section~\ref{sec:annotation}) which bounds the systematic component of style variation at an
intraclass correlation of $0.030$ but cannot speak to per-image disagreement. Annotation effort
was also concentrated---4 of 30 annotators produced 52\% of the corpus---so a systematic bias
shared by those four would not be visible as between-annotator variance, and one annotator
contributing $2\%$ of teeth does trace roots $10\%$ longer than consensus.

Primary and supernumerary teeth were excluded by protocol. The FDI uniqueness property that
motivates our semantic-segmentation formulation therefore does not hold for the estimated
1--3\% of patients presenting supernumerary teeth, and a deployed system must detect and defer
on that case.

Our cross-dataset evaluation confounds domain shift with annotation-protocol differences, since
DENTEX was annotated under its own guidelines. Some portion of the observed 62\% reduction
reflects disagreement about where a tooth boundary lies rather than model failure; separating
the two would require re-annotating a DENTEX subset under our protocol.

Finally, exhaustive correlation identified one patient imaged twice within our corpus. That pair
was constrained to a single cross-validation fold, but without patient identifiers we cannot
exclude further undetected repeats.

\section{Conclusion}
We assembled a densely annotated corpus of 1{,}422 panoramic radiographs with 42{,}142 tooth
polygons and used it to ask which factors govern performance on FDI tooth segmentation, holding
the corpus, evaluation code and post-processing fixed while varying one factor at a time.

Input resolution is the dominant lever, and it acts specifically on boundary precision:
mask mAP50-95 rises with resolution while mAP50 remains flat. Architecture is nearly irrelevant
in-domain, where a transformer with twice the parameters ties a one-stage detector, and matters
only marginally under domain shift. Three targeted interventions---a self-supervised encoder, a
promptable segmenter, and optimal anatomical label assignment---fail to meaningfully improve on a
conventional detector, the last despite correcting a constraint violated in 40\% of
out-of-domain predictions.

Taken together these results identify boundary precision as the binding constraint in panoramic
tooth segmentation. Detection and identification transfer robustly across acquisition
conditions; boundary placement does not. We therefore suggest that progress on this task will
come from resolution, acquisition diversity and annotation consistency rather than from
architectural novelty, and that reported comparisons should include an estimate of run-to-run
variance so that differences of the magnitude typically claimed can be distinguished from noise.

\section*{Ethics}
No institutional review board determination was sought for this study. The work is a
secondary analysis of panoramic radiographs already in the public domain, obtained from a
public redistribution rather than from a clinical archive, and was conducted without patient
contact, without intervention and without access to clinical records.

The imagery as obtained was not de-identified: filenames carried patient names and
acquisition timestamps. We stripped these and reassigned sequential identifiers before any
annotation or model training, so no annotator, model or reported result was exposed to an
identifier. The mapping from original filename to reassigned identifier is retained offline,
is not part of the corpus, and is not released. Image pixel data carries no embedded metadata.
The radiographs as analysed therefore carry no patient identifiers, no acquisition metadata
and no linkage to clinical records, and no attempt was made to re-identify individuals. The
annotation layer contributed by this work records anatomical structure only and introduces no
patient-level information.

We note that the upstream clinical origin of the imagery is not documented by the
redistributor, so the conditions under which it was originally acquired cannot be evidenced by
us.

\section*{Declaration of competing interest}
All authors are employed by AI Dentify, which develops commercial software for dental
radiographic analysis. The tooth segmentation model evaluated here is deployed in that company's
product, and the annotated corpus was produced by the company for the purpose of developing it.
The authors declare no other financial or personal relationships that could have appeared to
influence the work reported in this paper.
No external funding was received for this work.

\section*{CRediT authorship contribution statement}
\textbf{Muhammad Rehan:} Conceptualization, Software, Formal analysis, Investigation,
Visualization, Writing --- original draft, Writing --- review \& editing.
\textbf{Moaz Amjad:} Conceptualization, Methodology, Software, Investigation,
Writing --- original draft, Project administration. \textbf{Syed Danial Ahmed:} Validation, Data
curation. \textbf{Mariam Adnan:} Validation, Data curation.
\textbf{Haider Ali:} Writing --- original draft, Writing --- review \& editing.

\section*{Data availability}
The radiographs analysed here derive from a public redistribution whose upstream clinical
origin and licence terms are not documented by the redistributor; we therefore cannot
redistribute the imagery. We also do not cite the specific hosted record, because its filenames
contain patient names and acquisition dates and a citation would direct readers to identifiable
data; we will provide the source to the editors on request. The annotation layer is not
publicly released. The external cohort used for zero-shot evaluation (DENTEX) is publicly
available from its original authors under CC BY-NC-SA 4.0.

\section*{Code availability}
Code for training, evaluation, cross-validation and the contamination audit is released
at \url{https://github.com/Rehan000/opg-tooth-segmentation}.
This includes the duplicate detector based on exhaustive normalised correlation, and the
evaluation harness used to score all architectures on identical metrics. The DENTEX evaluation
is fully reproducible from public data with this code.

\clearpage

\bibliographystyle{cas-model2-names}
\bibliography{refs}

\appendix
\section{Per-class results}
\label{app:perclass}
Table~\ref{tab:perclass} gives the per-class breakdown underlying Fig.~\ref{fig:size} and the
aggregate metrics reported throughout. Support is listed separately for each evaluation set,
since a low score on a class present in few images carries less weight than the same score on a
well-represented one. Corpus prevalence is reported over all 1{,}422 annotated images and is
itself informative: third molars are absent or extracted far more often than any other class
(829 and 824 occurrences for teeth 18 and 28, against 1{,}404 for tooth 11), which is why they
combine large size with weak accuracy.

\begin{table*}[pos=t]\centering
\caption{Per-class results for the 1280\,px YOLO11m-seg model. Prevalence and mean annotated area are computed over all 1{,}422 annotated images; accuracy columns use the common evaluation harness. $\Delta$ is the zero-shot minus in-domain change in AP50-95. Classes are ordered by FDI quadrant.}
\label{tab:perclass}
\begin{tabular}{llrrrrrrrrr}
\toprule
 & & \multicolumn{2}{c}{Corpus} & \multicolumn{3}{c}{In-domain (val)} & \multicolumn{3}{c}{Zero-shot (DENTEX)} & \\
\cmidrule(lr){3-4}\cmidrule(lr){5-7}\cmidrule(lr){8-10}
FDI & Tooth & $n$ & Area & $n$ & IoU & AP & $n$ & IoU & AP & $\Delta$AP \\
\midrule
18 & UR 3rd molar & 829 & 4.0 & 80 & 0.782 & 0.681 & 341 & 0.484 & 0.224 & -0.457 \\
17 & UR 2nd molar & 1367 & 4.9 & 136 & 0.858 & 0.751 & 576 & 0.532 & 0.256 & -0.495 \\
16 & UR 1st molar & 1347 & 5.4 & 139 & 0.860 & 0.744 & 563 & 0.511 & 0.258 & -0.486 \\
15 & UR 2nd premolar & 1344 & 3.6 & 139 & 0.840 & 0.737 & 563 & 0.470 & 0.218 & -0.519 \\
14 & UR 1st premolar & 1279 & 4.0 & 125 & 0.814 & 0.690 & 586 & 0.446 & 0.164 & -0.526 \\
13 & UR canine & 1402 & 4.1 & 143 & 0.853 & 0.728 & 617 & 0.574 & 0.257 & -0.471 \\
12 & UR lateral incisor & 1391 & 3.2 & 143 & 0.860 & 0.727 & 622 & 0.575 & 0.231 & -0.496 \\
11 & UR central incisor & 1404 & 4.0 & 144 & 0.870 & 0.754 & 627 & 0.634 & 0.307 & -0.446 \\
\addlinespace
21 & UL central incisor & 1401 & 4.0 & 143 & 0.856 & 0.726 & 627 & 0.643 & 0.302 & -0.424 \\
22 & UL lateral incisor & 1386 & 3.1 & 141 & 0.845 & 0.720 & 616 & 0.597 & 0.247 & -0.473 \\
23 & UL canine & 1389 & 4.2 & 143 & 0.875 & 0.770 & 619 & 0.621 & 0.312 & -0.457 \\
24 & UL 1st premolar & 1287 & 4.1 & 132 & 0.801 & 0.682 & 575 & 0.522 & 0.212 & -0.470 \\
25 & UL 2nd premolar & 1338 & 3.7 & 135 & 0.802 & 0.702 & 559 & 0.501 & 0.231 & -0.471 \\
26 & UL 1st molar & 1346 & 5.5 & 132 & 0.844 & 0.718 & 544 & 0.565 & 0.294 & -0.424 \\
27 & UL 2nd molar & 1361 & 4.9 & 135 & 0.850 & 0.748 & 576 & 0.575 & 0.315 & -0.432 \\
28 & UL 3rd molar & 824 & 4.1 & 76 & 0.758 & 0.669 & 367 & 0.532 & 0.260 & -0.409 \\
\addlinespace
38 & LL 3rd molar & 1136 & 5.2 & 112 & 0.845 & 0.754 & 394 & 0.589 & 0.319 & -0.436 \\
37 & LL 2nd molar & 1357 & 6.0 & 139 & 0.880 & 0.784 & 577 & 0.613 & 0.330 & -0.454 \\
36 & LL 1st molar & 1333 & 6.2 & 136 & 0.866 & 0.764 & 473 & 0.596 & 0.290 & -0.474 \\
35 & LL 2nd premolar & 1364 & 3.6 & 138 & 0.848 & 0.728 & 587 & 0.591 & 0.288 & -0.440 \\
34 & LL 1st premolar & 1327 & 3.3 & 133 & 0.822 & 0.708 & 618 & 0.593 & 0.274 & -0.435 \\
33 & LL canine & 1416 & 3.5 & 142 & 0.849 & 0.730 & 629 & 0.641 & 0.322 & -0.408 \\
32 & LL lateral incisor & 1392 & 2.3 & 138 & 0.799 & 0.644 & 629 & 0.607 & 0.276 & -0.368 \\
31 & LL central incisor & 1395 & 1.9 & 137 & 0.767 & 0.586 & 625 & 0.594 & 0.231 & -0.354 \\
\addlinespace
48 & LR 3rd molar & 1140 & 5.1 & 120 & 0.851 & 0.734 & 402 & 0.582 & 0.321 & -0.412 \\
47 & LR 2nd molar & 1347 & 6.0 & 135 & 0.872 & 0.765 & 581 & 0.604 & 0.351 & -0.413 \\
46 & LR 1st molar & 1339 & 6.2 & 136 & 0.856 & 0.752 & 481 & 0.569 & 0.276 & -0.476 \\
45 & LR 2nd premolar & 1363 & 3.6 & 135 & 0.835 & 0.745 & 585 & 0.599 & 0.305 & -0.439 \\
44 & LR 1st premolar & 1337 & 3.4 & 134 & 0.821 & 0.704 & 619 & 0.589 & 0.281 & -0.423 \\
43 & LR canine & 1414 & 3.5 & 142 & 0.850 & 0.716 & 629 & 0.655 & 0.354 & -0.362 \\
42 & LR lateral incisor & 1390 & 2.3 & 137 & 0.818 & 0.676 & 627 & 0.605 & 0.252 & -0.424 \\
41 & LR central incisor & 1397 & 2.0 & 140 & 0.776 & 0.595 & 624 & 0.567 & 0.208 & -0.387 \\
\midrule
\multicolumn{2}{l}{Mean} & & & & 0.835 & 0.717 & & 0.574 & 0.274 & -0.443 \\
\bottomrule
\end{tabular}
\end{table*}

\clearpage
\addtocounter{page}{-1}

\end{document}